\documentclass[runningheads]{llncs}
\usepackage{graphicx}
\usepackage{caption} 
\usepackage{subcaption}
\usepackage{amsmath,amssymb}
\usepackage{bm}
\usepackage{bbm}
\usepackage{booktabs}
\begin{document}
\title{Class-Similarity Based Label Smoothing for Confidence Calibration}
%
%
\author{Chihuang Liu\inst{*} \and Joseph JaJa}
\authorrunning{C. Liu \and J. JaJa}
%
\institute{University of Maryland, College Park MD 20742, USA\\
\email{chliu@umd.edu}}

\maketitle

\begin{abstract}
Generating confidence calibrated outputs is of utmost importance for the applications of deep neural networks in safety-critical decision-making systems. The output of a neural network is a probability distribution where the scores are estimated confidences of the input belonging to the corresponding classes, and hence they represent a complete estimate of the output likelihood relative to all classes. In this paper, we propose a novel form of label smoothing to improve confidence calibration. Since different classes are of different intrinsic similarities, more similar classes should result in closer probability values in the final output. This motivates the development of a new smooth label where the label values are based on similarities with the reference class. We adopt different similarity measurements, including those that capture feature-based similarities or semantic similarity. We demonstrate through extensive experiments, on various datasets and network architectures, that our approach consistently outperforms state-of-the-art calibration techniques including uniform label smoothing.
\end{abstract}

\section{Introduction}\label{section_introduction}

Machine learning algorithms have progressed rapidly in recent years and are becoming the critical component in a wide variety of technologies~\cite{goodfellow2016deep}. In most of these applications, making wrong decisions could lead to very high costs, including significant business losses or even severe human injuries~\cite{brundage2018malicious}. As a result, in real-world decision-making systems, machine learning models should not only try to be as accurate as possible, but should also indicate when they are likely to be incorrect, which allows the decision-making to be stopped or passed to human experts when the models are not sufficiently confident to produce a correct prediction. It is therefore strongly desirable that a network provides a calibrated confidence measure in addition to its prediction; that is, the probability associated with the predicted class label should reflect its ground truth likelihood of correctness~\cite{naeini2015obtaining}. 

However, recent works~\cite{guo2017calibration} have shown through extensive empirical studies that even though impressively accurate, modern deep neural networks are poorly calibrated. It turns out that modern DNNs are overconfident - the prediction accuracy is likely to be lower than what is indicated by the associated confidence. Since the discovery of this challenging problem, several methods ~\cite{zhang2020mix,muller2019does,xing2019distance,thulasidasan2019mixup} have been explored and empirically shown to improve confidence calibration performance on the predictions, which we refer to as \textit{prediction calibration}, for which only the model's prediction (the winning class) and its associated confidence (the maximum softmax score) are considered.

In the output of a neural network, each score represents the model's estimated probability of the current input belonging to that corresponding class. Therefore, for a well-calibrated model, each probability should be indicative of the actual likelihood of the input belonging to each class, not just the one with the maximum score~\cite{kumar2019verified,kull2019beyond}. This is significant in safety-critical applications. For example, an autonomous driving system predicts an object to be 60\% a pedestrian, 30\% an obstacle, and 10\% a traffic sign. Clearly these probabilities should be calibrated. We refer to this as \textit{output calibration} where the entire output probabilities are considered.

The output scores define a probability distribution over all classes, and we call the perfectly calibrated distribution as the optimal output distribution. Under the optimization scheme with cross-entropy loss that most modern DNNs adopt, a model would achieve perfect confidence calibration if directly trained using the optimal distribution as training label. Unfortunately, as one would expect, it is unclear how to directly compute the optimal distribution in general. However, under certain reasonable assumptions, we can develop good approximations. Different classes are not equally distinct, and a class can be inherently more similar to some classes than others. For example, in the CIFAR-100 dataset, we could generally agree that the class dolphin is much more similar to the class seal than to the class rose, and the probability of a seal should be higher than a rose in the output distribution of a dolphin input. Therefore, we make the following two assumptions:
\begin{enumerate}
    \item The distribution of other (i.e. non ground-truth) classes is non-uniform in general.
    \item The probability value should correlate positively with the similarity between the true class and any other class, i.e. the more similar they are the higher the value.
\end{enumerate}
In this paper, we propose a novel form of the smooth labeling, called \textit{class-similarity based label smoothing}, which uses class similarities to approximate the optimal output distribution. In the proposed smooth label of a reference class, the label for another class is based on its similarity with the reference class, and hence more similar classes result in higher values. 

From the label smoothing perspective, our proposed smooth label is more intuitive than the traditional uniform smooth label. One-hot labels have all probability mass in one class, which are zero-entropy signals that admit no uncertainty about the input. When a network is trained using such labels, it inevitably becomes overconfident. With uniform smooth labels, the output of a network is trained to be a mixture of a Dirac distribution and a uniform distribution. This implies that the predicted probabilities for other classes are encouraged to be equal. As we discussed above, the label value should be based on the similarity between classes, which is not accounted for in either the one-hot or the uniform smooth labels. Relational information can be crucial and provides high-order properties that can improve performance of a model in various tasks~\cite{park2019relational}.

In our proposed method, we first measure the similarity between classes which is then mixed with the one-hot labels to serve as the final smooth label. Since measuring similarity is in general an open problem~\cite{wang2004image,zhang2018unreasonable}, we adopt different metrics, including a new notion of semantic similarity, and evaluate their effectiveness in confidence calibration on various data benchmarks and architectures.

\section{Related Work}
Since the pioneering work~\cite{guo2017calibration}, confidence calibration has raised a great interest in the machine learning community resulting in a great deal of recent work trying to address this issue. Existing calibration techniques can be broadly categorized based on whether or not they are post-hoc methods.

Post-hoc calibration methods use parametric or non-parametric models to transform the network's predictions based on a held-out validation set to improve calibration. Traditional techniques include Platt scaling~\cite{platt1999probabilistic}, Isotonic Regression~\cite{zadrozny2001obtaining}, Bayesian binning~\cite{naeini2015obtaining,zadrozny2001obtaining}, Histogram binning~\cite{zadrozny2001obtaining}, matrix and vector scaling~\cite{guo2017calibration}, Beta calibration~\cite{kull2017beta}, and temperature scaling~\cite{guo2017calibration}. Among these methods, temperature scaling consistently outperforms the other methods~\cite{guo2017calibration}. More recent advances include Dirichlet calibration~\cite{kull2019beyond}, ensemble temperature scaling~\cite{zhang2020mix}, mutual information based binning~\cite{patel2020multi}, and spline recalibration~\cite{gupta2020calibration}. Non-post-hoc methods are mostly based on adapting the training procedure, including modifying the training loss~\cite{mukhoti2020calibrating,xing2019distance}, label smoothing~\cite{szegedy2016rethinking,muller2019does}, and data augmentation~\cite{thulasidasan2019mixup,yun2019cutmix}. 

Another line of related work is Bayesian Neural Networks~\cite{gal2016dropout,maddox2019simple,ritter2018scalable}. Bayesian methods provide a natural probabilistic representation of uncertainty in deep learning and are well-suited for providing calibrated uncertainty estimation. However, these methods are resource-demanding and hard to scale to modern datasets and architectures~\cite{maddox2019simple}.


\section{Confidence Calibration}
\label{section_confidence_calibration}
In this section, we formally introduce the definition of confidence calibration. We denote the input as $X\in\mathcal X$ and label as $Y\in\mathcal Y=\{1,2,...,K\}$. Let $h$ be a neural network and $h(k|x)$ is the confidence estimate of the sample $x$ belonging to class $k$. The prediction is the winning class $\hat Y=\arg\max h(Y|X)$ and its associated confidence is the maximum $\hat P=\max h(Y|X)$. The number of samples is $n$.

\subsection{Prediction and Output Calibration}
\label{subsection_prediction_calibration}
In prediction calibration, only the winning class and its associated confidence are considered. With perfect prediction calibration, the confidence estimate $\hat P$ represents a true probability that indicates the likelihood of correctness
\begin{equation}\label{equation_prediction}
\mathbb P(Y=\hat Y\,|\,\hat P=p) = p, \,\,\,\forall p\in[0, 1]
\end{equation}

The output of a neural network $h(Y|X)$ is the confidence estimate for all classes. For a calibrated model, we would like not only the maximum but all confidences to be calibrated, which means that $h(k|X)$ represents the actual likelihood of $Y=k$ for all classes $k$. Perfect output calibration is defined as
\begin{equation}\label{equation_output}
  \begin{aligned}
\mathbb P\left(Y=k\,|\,h(k|X)=p\right) = p, \,\,\, & \forall p\in[0, 1]\,\,\text{and}\,\,\forall k\in\{1,2,...,K\}
\end{aligned}  
\end{equation}
Note that output calibration infers prediction calibration but not vice versa.

\subsection{Histogram-Based ECE}
Since the probabilities in (\ref{equation_prediction}) and (\ref{equation_output}) cannot be computed using finitely many samples, empirical approximations are developed. The most widely adopted estimator is histogram-based expected calibration error (ECE)~\cite{naeini2015obtaining,guo2017calibration}.

The model's output probabilities $p_i^k=h(k|x_i)$ are grouped into $M$ interval bins, where $p_i^k$ is the model's output confidence for sample $i$ belonging to class $k$. Let $B_m$ be the set of indices of which the confidence falls into bin $m$, i.e. $p_i^k\in (\frac{m-1}{M}, \frac{m}{M}]$. The likelihood and confidence of $B_m$ are defined as
\begin{equation}
\begin{aligned}
\text{lik}(B_m) & = \frac{1}{|B_m|}\sum_{(i, k)\in B_m}\mathbbm{1}(y_i = k)\quad\text{and} \quad
\text{conf}(B_m) & = \frac{1}{|B_m|}\sum_{(i, k)\in B_m} p_i^k
\end{aligned}
\end{equation}
where $y_i$ is true class label for sample $i$. The histogram-based ECE is defined as
\begin{equation}
\text{Histogram ECE} = \sum_{m=1}^M\frac{|B_m|}{nK}|\text{lik}(B_m)-\text{conf}(B_m)|
\end{equation}

\subsection{KDE-Based ECE}
While a histogram-based estimator is easy to implement, it inevitably inherits drawbacks from histograms, for example being sensitive to the binning schemes and data-inefficient~\cite{kumar2019verified}. KDE-based estimator is proposed in \cite{zhang2020mix} by replacing histograms with density estimation using a continuous kernel. 

Let $\phi: \mathbb R\rightarrow\mathbb R_{\geq 0}$ denote a kernel function and $h$ denote the bandwidth. The density function and  canonical calibration function are given by 
\begin{equation}
    \begin{aligned}
    f(p) & = \frac{h^{-K}}{n}\sum_{i=1}^n\prod_{k=1}^K\phi(\frac{p-p_i^k}{h}) \quad\text{and} \quad
    c(p) & = \frac{\sum_{i=1}^ny_i\prod_{k=1}^K\phi(\frac{p-p_i^k}{h})}{\sum_{i=1}^n\prod_{k=1}^K\phi(\frac{p-p_i^k}{h})}
    \end{aligned}
\end{equation}
and the KDE-based ECE is computed as
\begin{equation}
\text{KDE ECE} = \int \|p-c(p)\|f(p)dp
\end{equation}
While KDE alleviates the dependence on histogram binning, it heavily depends on the kernel choice and may induce error from the integral approximation procedure. Therefore, we use both ECE metrics in our evaluation. 

\section{Learning with Different Labels}
\label{section_learn_with_label}
In this section, we  discuss the learning objective with respect to different labels and present the intuition behind our method. The output of a neural network is a probability distribution over all classes
$h(k|x)=\exp(z_k)/\sum_{i=1}^K\exp(z_i)$, 
where $z_k$ is the activation in the last layer. Let $\pi(k|x)$ be the label corresponding to input $x$. The model is trained by minimizing the cross-entropy loss
\begin{equation}
l(x)=-\textstyle\sum_{i=1}^K\pi(k|x)\log(h(k|x))
\end{equation}
With one-hot label, the cross entropy loss can be simplified to
\begin{equation}
l(x)=-\log(h(y|x))
\end{equation}
Under this loss, the model is not only trained to make a correct prediction but also with the highest confidence $z_y$ possible to reduce the loss, which causes the model to become overconfident. Uniform smooth labels are defined by 
\begin{equation}
\pi(k|x)=(1-\alpha)e_y+\alpha u(k)/K
\end{equation}
where $e_y$ is the coordinate vector and $u$ is a uniform distribution. The cross-entropy loss with uniform smooth label can be written as
\begin{equation}
l(x)=-(1-\alpha)\log(h(y|x))+\alpha H(u, h)
\end{equation}
The second part of this loss encourages the model to match its output with a uniform distribution, which implies that the example is regarded to be equally probable in any other class. This is clearly not the case in general as some classes are inherently more similar than others. The intuition behind our method is that we assume there is an unknown optimal distribution for perfect calibration $q^{*}(k|x)$ that satisfies the two assumptions made in Section~\ref{section_introduction} and can be approximated by estimating the class similarities. Let $\hat q\approx q^{*}$ be the approximated optimal distribution, then our proposed smooth label is defined as
\begin{equation}
\pi(k|x)=(1-\alpha)e_y+\alpha\hat q(k|x)
\end{equation}
With this label, the cross entropy loss can be written as
\begin{equation}
l(x)=-(1-\alpha)\log(h(y|x))+\alpha H(\hat q, h)
\end{equation}
and the model is trained to make a correct classification while matching it output with the approximated optimal probability distribution. Hence, \textit{minimizing $H(\hat q, h)$ is a direct optimization for both prediction and output calibration}.

\section{Approach}
\label{section_approach}
In this section, we describe our approach to capture class similarities in order to compute the approximation distribution $\hat q$ and generate our proposed smooth labels. Capturing semantic similarity has been a longstanding and still a wide-open problem~\cite{wang2004image,zhang2018unreasonable}. Different metrics quantify the similarity from different perspectives, therefore we propose to use several distance metrics, including a novel one based on word2vec mapping of label words, and evaluate the performance under varying notions of the captured similarities. 

\subsection{Image Space}
For directly computing the distance in the image space, we use $L_p$ norms and choose $p=1$ and $p=2$. For two inputs $x$ in class $k$ and $x'$ in class $k'$, the pairwise distances are given by

\textbf{$\bm{L_p}$ Distance}
\begin{equation}
d(x, x') = \|x-x'\|_p
\end{equation}

\subsection{Representation Space}
Studies have shown that features learned by neural networks are often surprisingly useful as a representational space for a much wider variety of tasks and match with human perception~\cite{dosovitskiy2016generating,zhang2018unreasonable}. The latent representations of autoencoders have been shown to be useful for various downstream tasks~\cite{bengio2013representation}. Therefore, we also propose to use an autoencoder to map the data to the representation space and compute distances between their encodings. 

\textbf{Autoencoder Distance}
\begin{equation}
d(x, x') = \|r(x)-r(x')\|_2
\end{equation}
where $r(x)$ is the autoencoder latent encoding of input $x$.

The inter-class distance between classes $k$ and $k'$ is determined by averaging the distances between all pairs of inputs that belong to them
\begin{equation}
d_k(k') = \frac{1}{|C(k)||C(k')|}\sum_{x\in C(k), x'\in C(k')}d(x, x')
\end{equation}
where $C(k)$ is the set of data points in class $k$, and $d_k$ is a vector that contains the distances between class $k$ and all other classes.

\subsection{Semantic Space}
The previous distance metrics use either the original features of the objects or those generated by a neural network. In general, samples in the same class are expected to share a common semantic meaning, which may not be captured in either feature space. It was observed for example that image visual similarity is not necessarily the same as semantic similarity~\cite{deselaers2011visual}. The problem of defining semantic similarity has been studied extensively in the NLP literature as well as in various disciplines for which knowledge can be captured through an ontology or a hierarchy of classes~\cite{lee2008comparison}. For most datasets, the words used to label each class capture significant semantics of the class~\cite{miller1995wordnet}. We make use of the advances in NLP based on the labels to define the semantic similarity. Vector representation of words has been shown to successfully capture semantic similarities such that words with similar meaning are mapped to similar points in the vector space~\cite{mikolov2013efficient}. Therefore, we propose to use a word2vec model to map the label words into Euclidean space, and then compute the distances between the vectors. 

\textbf{Word Embedding Distance}
\begin{equation}
d_k(k') = \|\mathcal V(w(k))-\mathcal V(w(k'))\|_2
\end{equation}
where $\mathcal V$ is a word2vec model and $w(k)$ be the natural language word associated with class $k$. This notion can be generalized to the case in which each class is defined by a set of words or a sentence.

\subsection{Class-Similarity Based Smooth Label}
Class distances are converted into class similarities using the softmax function
\begin{equation}
s_k(k') = \frac{\exp(-\beta d_k(k'))}{\sum_{i=1}^K\exp(-\beta d_k(i))}
\end{equation}
where $\beta\geq 0$ is a hyperparameter that controls how ``uniform'' the similarity distribution is. In order to ensure the consistency of $\beta$ for different metrics, we normalize the distances $d_k$ to zero mean and standard deviation of one before applying the softmax function. Note that this normalization is equivalent to scaling $\beta$ and does not affect the relative relationship between classes.

Finally, the class-similarity based smooth label is defined as follows. Let $e_k$ be the one-hot label vector, the smooth label for class $k$ is
\begin{equation}
y_k = (1-\alpha)e_k+\alpha s_k
\end{equation}
where $\alpha$ is the label smoothing factor. Note that in practice we set $s_k(k)=0$ and scale $\sum_{k'\neq k}s_k(k')=1$ to make sure that $\alpha$ consistently represents the total mass in the label over the other (non ground-truth) classes. Otherwise the total mass will be different because of different values of $s_k(k)$.

\section{Experiments}
\label{section_experiments}

\subsection{Setup}

We compare the four variants of our proposed method with $L_1$ distance, $L_2$ distance, Autoencoder distance (AE) and Word Embedding distance (WE) to the vanilla training using one-hot labels, as well as various techniques that improve confidence calibration: temperature scaling (TS)~\cite{guo2017calibration}, uniform label smoothing~\cite{szegedy2016rethinking,muller2019does}, mixup training~\cite{thulasidasan2019mixup}, Dirichlet calibration with
off-diagonal regularization (Dir-ODIR)~\cite{kull2019beyond}, and
ensemble temperature scaling (ETS)~\cite{zhang2020mix}.

We perform experiments on CIFAR-100 and Tiny-ImageNet using DenseNet (DN), ResNet (RN), and WideResNet(WRN). For all experiments, the network architectures and all parameters are identical for all methods. For CIFAR-100, we use DenseNet-161, ResNet-18, and WRN with $\{16, 16, 32, 64\}$ filters respectively. For Tiny-ImageNet, we use DenseNet-161, ResNet-34, and WRN-50-2. The dimensionality of the latent space of the autoencoder is set to 256 for CIFAR-100 and 1024 for Tiny-ImageNet. We use a pretrained Wikipedia2Vec~\cite{yamada2018wikipedia2vec} as our word2vec model $\mathcal V$ with a vector of length 100. For histogram-based ECE, we set the number of interval bins $M=15$ following \cite{guo2017calibration,muller2019does}. For KDE-based ECE, we use the Triweight Kernel $K(u)=\frac{35}{32}(1-u^2)^3$ and bandwidth $h=1.06\sigma n^{-1/5}$~\cite{zhang2020mix}.

\subsection{Confidence Calibration Results}
\label{subsection_calibration_results}
\begin{table}[t]
\caption{Histogram and KDE-based ECE (\%) results of prediction (P) and output (O) on CIFAR-100.\label{table_ece_cifar100}}
    \centering
\resizebox{\textwidth}{!}{
    \begin{tabular}{cccccccccccc}
    \toprule
    Model &ECE & One-hot & TS & Uniform & Mixup & DirODIR & ETS &$L_1$(Ours) & $L_2$(Ours) & AE(Ours)  & WE(Ours)\\ \midrule
    DN  & Hist P & 16.40 & 2.08 & 2.41 & 4.58 & 2.10 & 2.04 & 2.39 & 2.65 & 1.96 & \bf 1.68\\
    DN  & Hist O & 35.52 & 3.55  & 8.11 & 8.81 & 3.54 & 3.48 & 4.15 & 4.74 & 3.32 & \bf 2.98\\
    DN  & KDE P & 15.28 & 1.90 & 2.68 & 4.52 & 1.95 & 1.88 & 2.61 & 2.72 & 2.08 & \bf 1.75\\
    DN  & KDE O & 28.65 & 7.01 & 15.08 & 13.40 & 7.09 & \bf 7.00 & 12.85 & 12.92 & 12.44 & 11.32\\ \midrule
    RN  & Hist P & 22.96 & 1.98 & 2.35 & 2.70 & 1.85 & \bf 1.71 & 2.84 & 2.76 & 2.02 & 1.74\\
    RN  & Hist O & 50.07 & 3.13 & 3.80 & 5.03 & 2.98 & 2.90 & 5.06 & 5.10 & 3.34 &\bf 2.47\\
    RN  & KDE P & 21.13 & 2.16 & 2.60 & 2.83 & 1.84 &\bf 1.65 & 2.91 & 2.76 & 2.19 & 1.76\\
    RN  & KDE O & 39.23 & 9.49 & 15.17 & 12.03 & 10.42 & 9.14 & 11.89 & 11.15 & 10.13 &\bf 9.02\\ \midrule
    WRN  & Hist P & 12.17 & 2.34 & 2.77 & 2.29 & 2.07 & 1.79 & 1.68 & 1.56 & 0.77 & \bf 0.72\\
    WRN  & Hist O & 25.71 & 5.11 & 5.41 & 5.02 & 4.54 & 3.91 & 2.91 & 2.60 & 1.64 & \bf 1.55\\
    WRN  & KDE P & 11.57 & 2.35 & 2.83 & 2.49 & 1.99 & 1.83 & 1.57 & 1.48 & 1.15 & \bf 1.11\\
    WRN  & KDE O & 21.13 & 9.08 & 9.27 & 9.47 & 6.87 & 6.92 & 7.59 & 7.01 &  4.53 & \bf 3.73\\
    \bottomrule
    \end{tabular}}
\end{table}

\begin{table}[t]
\scriptsize
\caption{Histogram and KDE-based ECE (\%) results of prediction (P) and output (O) on Tiny-ImageNet.\label{table_ece_tinyimagenet}}
    \centering
\resizebox{\textwidth}{!}{
    \begin{tabular}{cccccccccccc}
    \toprule
    Model &ECE & One-hot & TS & Uniform & Mixup & DirODIR & ETS &$L_1$(Ours) & $L_2$(Ours) & AE(Ours)  & WE(Ours)\\ \midrule
    DN  &Hist P & 13.98 & 3.45 & 3.77 & 3.70 & 3.23 & 2.16 & 2.93 & 3.07 & 2.54 & \bf 1.19\\
    DN  & Hist O & 29.76 & 11.64 & 13.27 & 12.13 & 10.57 & 8.80 & 9.94 & 10.40 &  7.74 & \bf 3.00\\
    DN  &KDE P & 13.38 & 3.46 & 3.72 & 3.76 & 3.21 & 2.21 & 2.95 & 3.13 &  2.60 & \bf 1.23\\
    DN  & KDE O & 27.64 & 18.36 & 21.01 & 19.06 & 17.87 & 17.48 & 17.51 & 18.18 & 17.56 & \bf 15.67\\ \midrule
    RN  & Hist P & 24.95 & 4.83 & 5.24 & 5.87 & 5.15 & 3.12 & 2.95 & 2.86 & 2.49 & \bf 1.23\\
    RN  & Hist O & 58.01 & 9.94 & 10.92 & 11.76 & 10.07 & 6.52 & 5.78 & 5.59 &5.05 & \bf 2.12\\
    RN  & KDE P & 23.55 & 4.93 & 5.21 & 5.80 & 5.17 & 3.01 & 3.02 & 2.82 & 2.57 & \bf 1.26\\
    RN  & KDE O & 46.51 & 22.33 & 27.57 & 24.99 & 23.14 & 13.15 & 27.21 & 28.60 & 29.97 & \bf 11.84\\ \midrule
    WRN  & Hist P & 22.39 & 10.15 & 12.92 & 8.73 & 9.79 & 6.71 & 8.55 & 8.03 & 7.41 &\bf 4.21\\
    WRN  & Hist O & 52.14 & 22.19 & 25.10 & 17.00 & 19.17 & 14.36 & 16.87 & 16.59 & 13.82 & \bf 7.93\\
    WRN  & KDE P & 21.37 & 10.75 & 12.90 & 8.74 & 9.78 & 6.55 & 8.50 & 7.97 & 7.45 & \bf 4.13\\
    WRN  & KDE O & 43.78 & 28.14 & 37.64 & 26.76 & 27.34 & 19.44 & 25.81 & 24.80 & 24.33 & \bf 11.30\\
    \bottomrule
    \end{tabular}}
\end{table}

We show confidence calibration results on CIFAR-100 in Table~\ref{table_ece_cifar100} and Tiny-ImageNet in Table~\ref{table_ece_tinyimagenet}. We set $\alpha=0.1$ and $\beta\in[0.5, 6]$. Parameter effects are discussed in detail in Section~\ref{subsection_parameters}.

The models trained with our proposed method generally outperforms other methods. Among the four distance measures, WE performs the best in all scenarios  which is expected considering that the distance in the vector space more faithfully reflects class semantic relationships than pixel-wise distances and the latent encoding distance. WE delivers better confidence calibration than all comparison methods in most cases. By comparing AE with WE, we can see the performance gap is more significant on Tiny-ImageNet than on CIFAR-100. As the complexity of a dataset and the number of classes increase, it becomes more difficult for an autoencoder to learn good representations, which leads to the performance difference. The large output ECE difference between WE and Uniform shows that the uniform distribution is not an optimal objective. We note that for output ECE, the KDE estimate is significantly distinct from the histogram estimate. The output probabilities of a model contain a majority of near-zero values, thus in this case the density estimation is not very accurate.

\begin{figure}[t]
\centering
\subcaptionbox{Prediction on CIFAR-100} {\includegraphics[width=.236\textwidth]{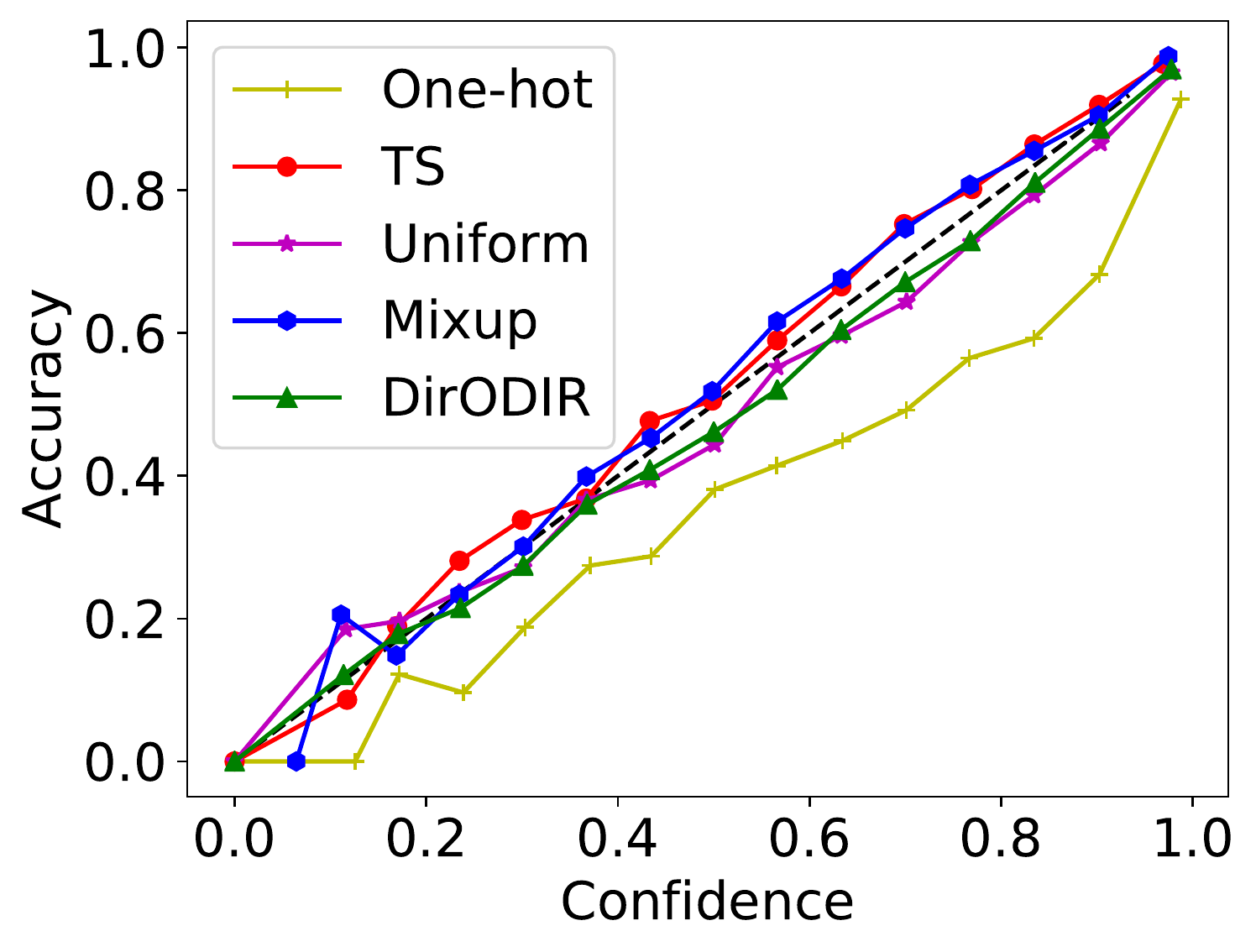}
  \includegraphics[width=.24\textwidth]{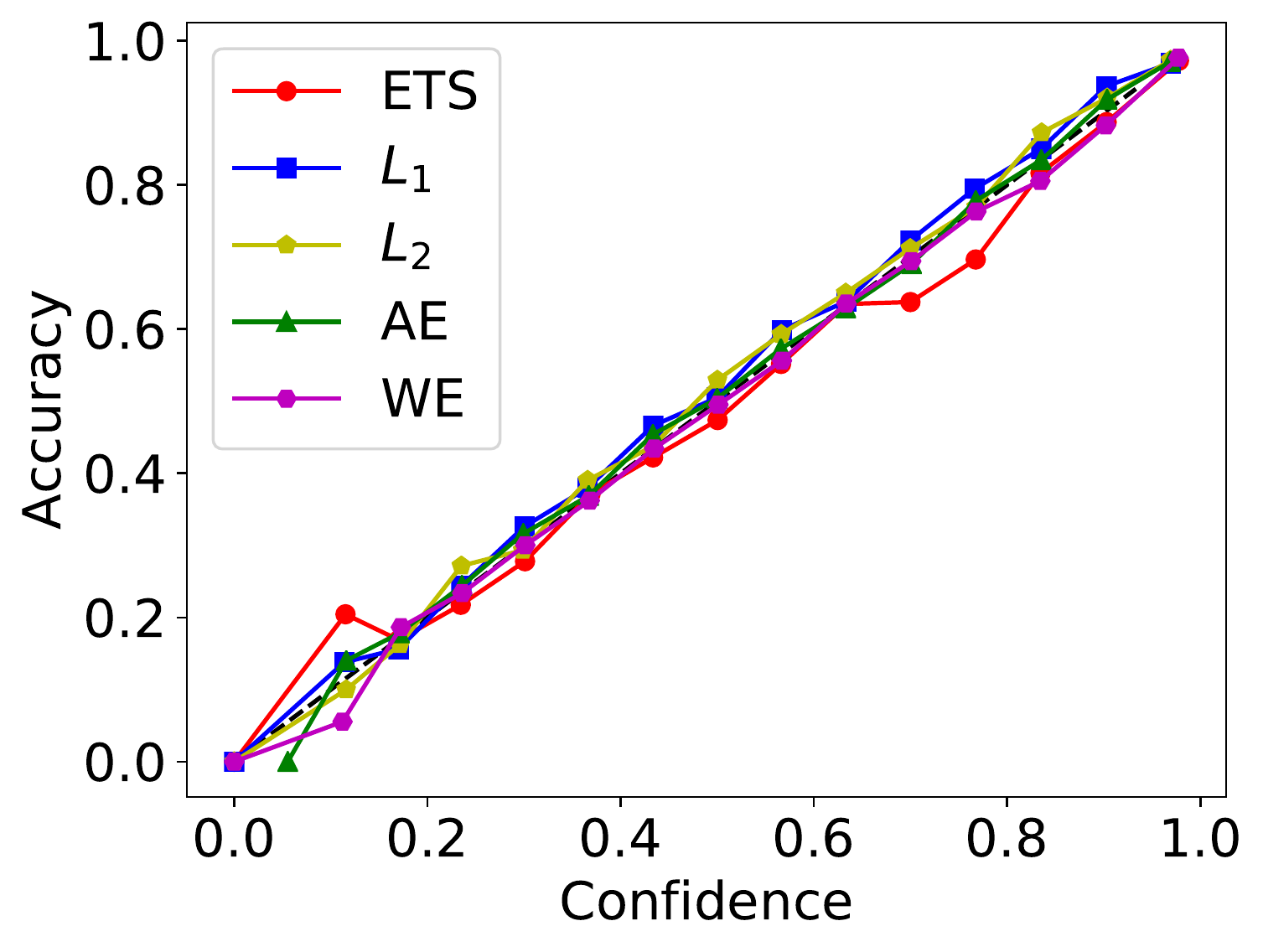}
  }%
\hfill
\subcaptionbox{Output on CIFAR-100} {\includegraphics[width=.236\textwidth]{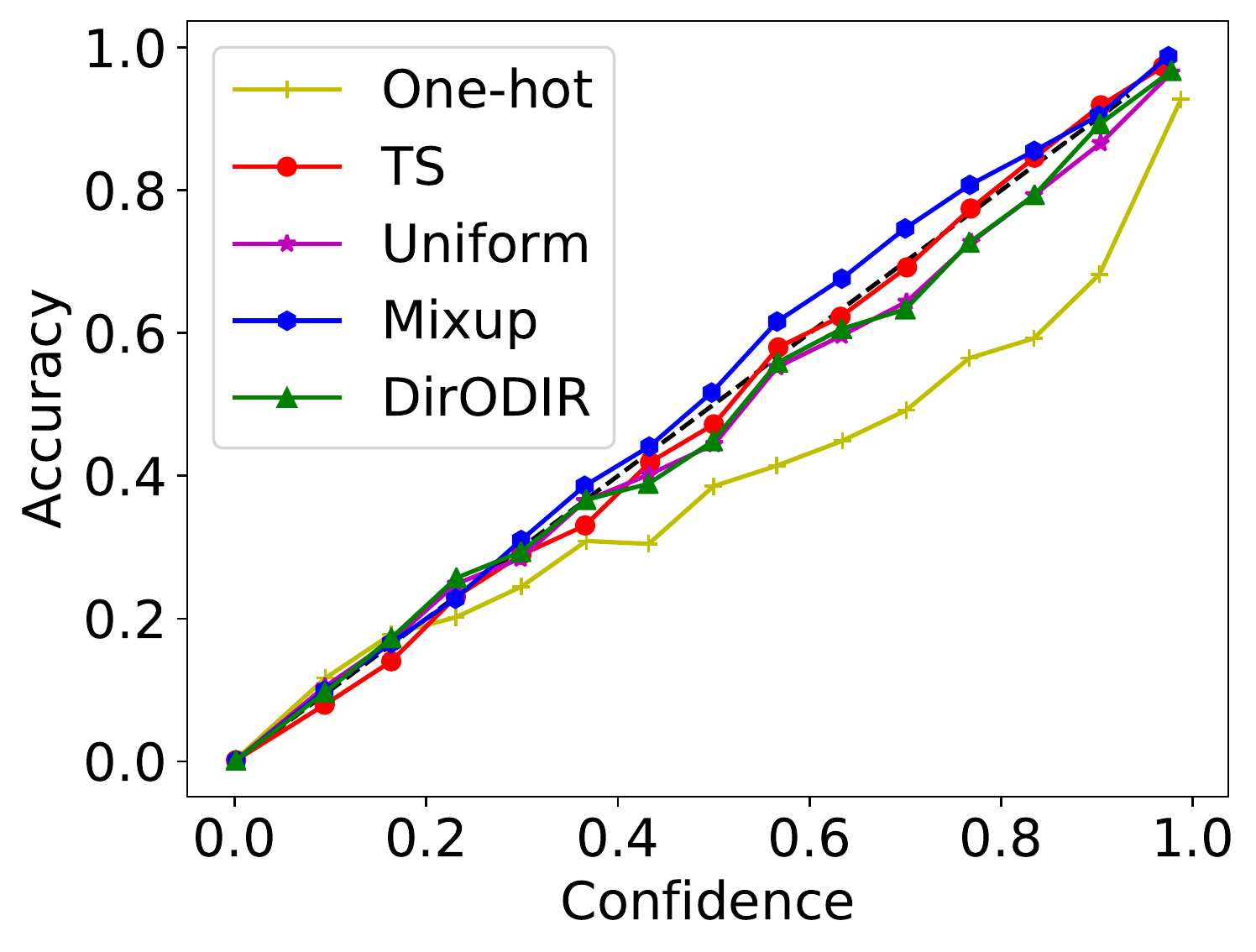}
  \includegraphics[width=.24\textwidth]{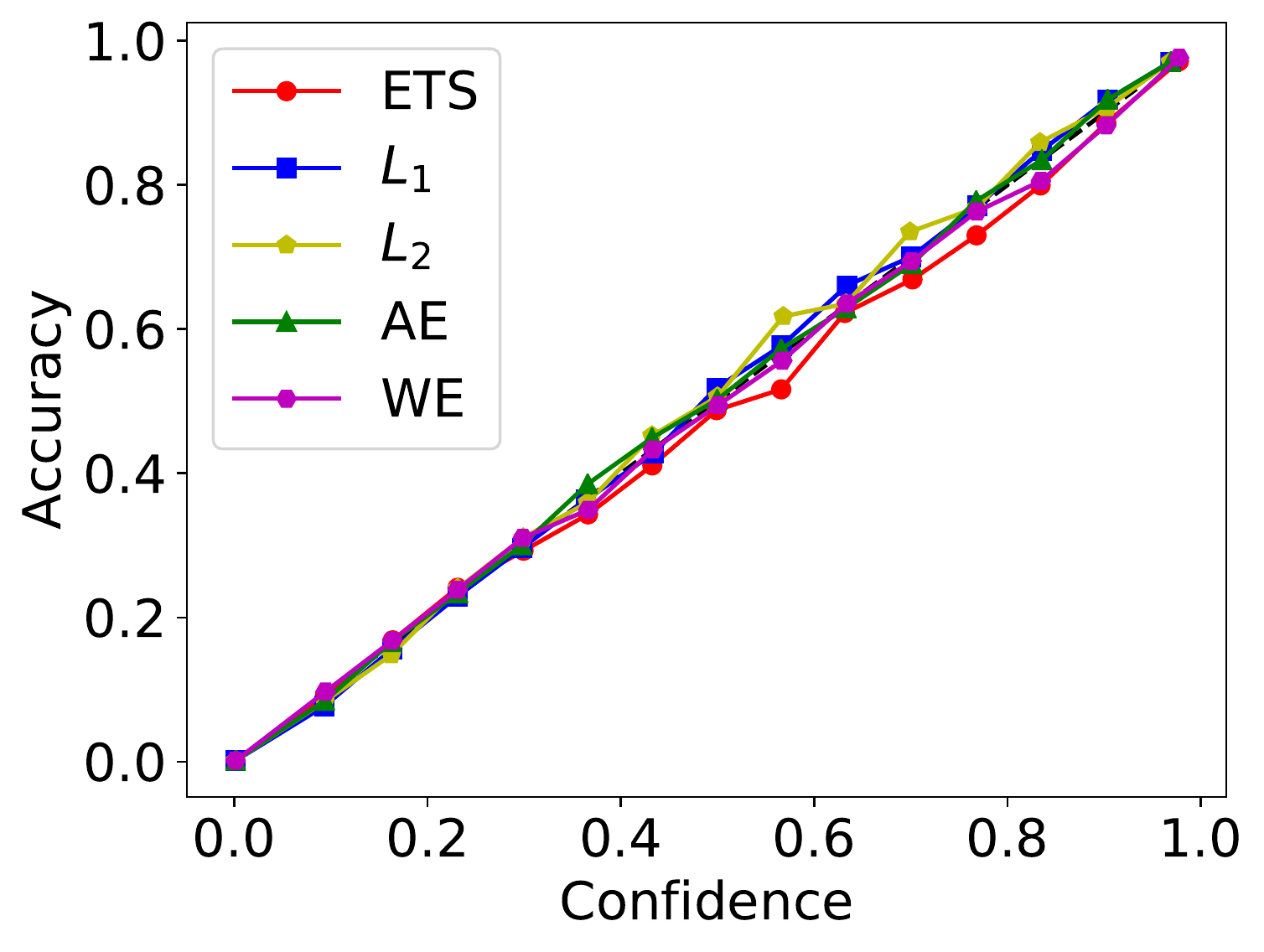}   }%

\subcaptionbox{Prediction on Tiny-ImageNet}{\includegraphics[width=.236\textwidth]{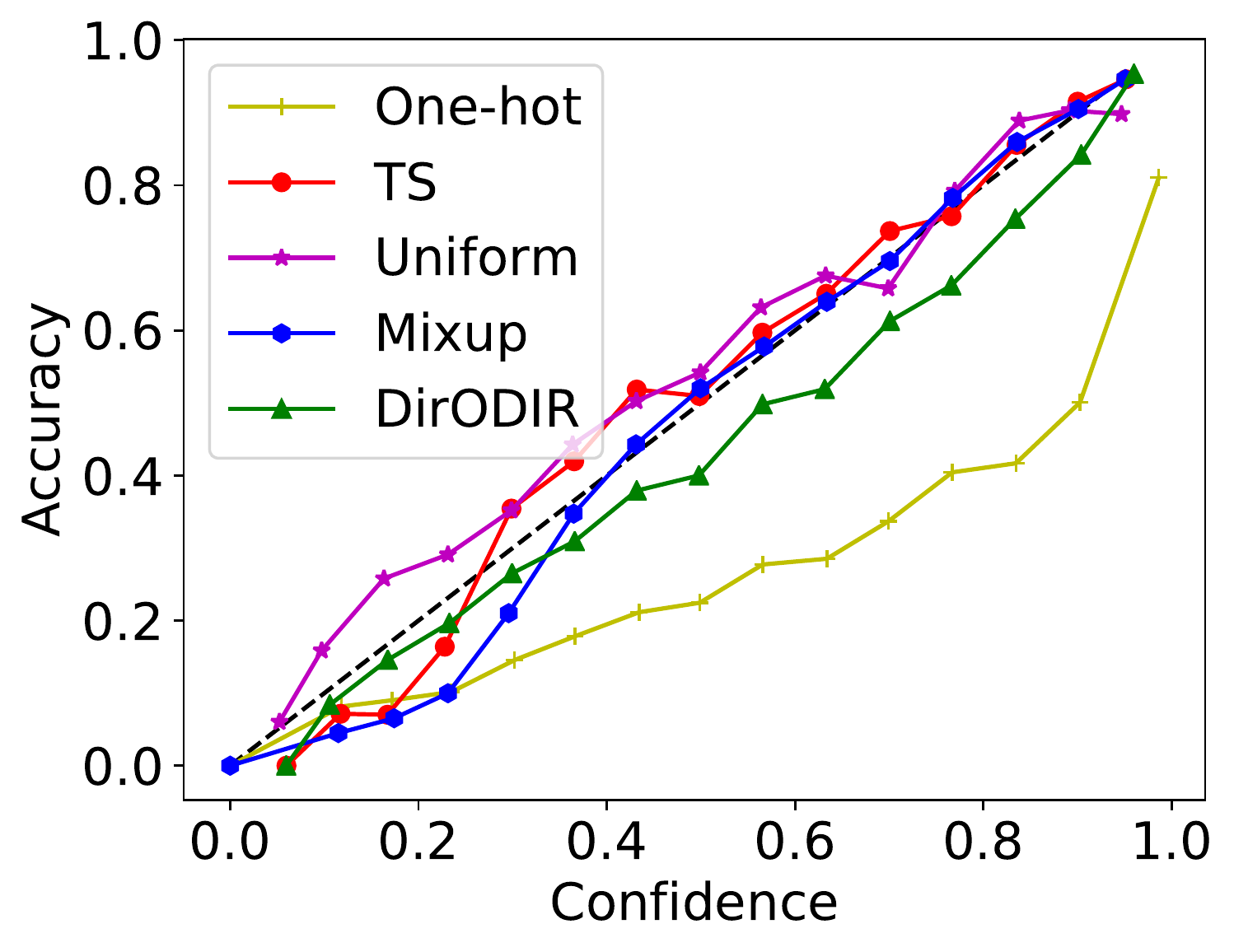}
  \includegraphics[width=.24\textwidth]{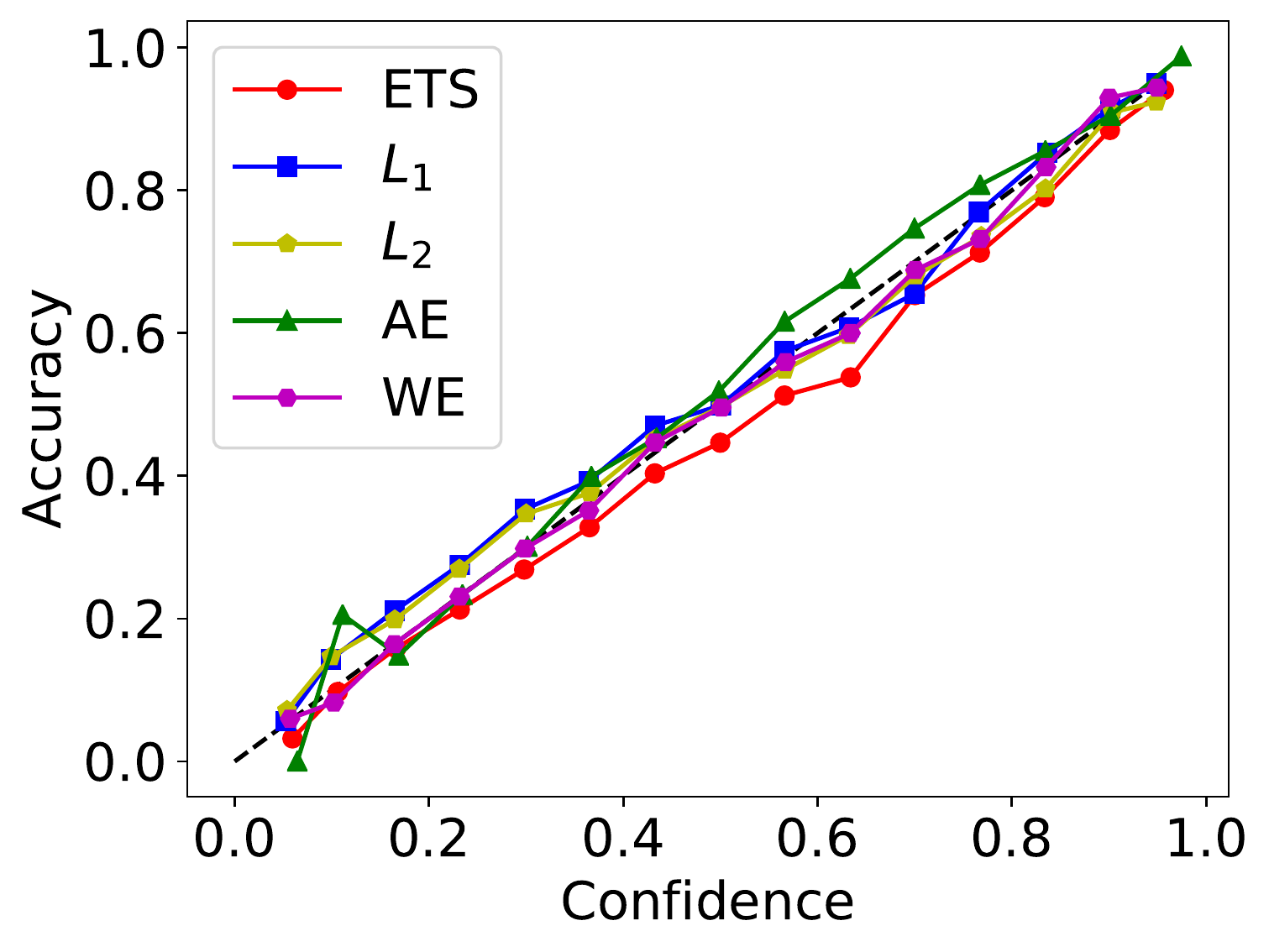}}%
\hfill
\subcaptionbox{Output on Tiny-ImageNet}
{\includegraphics[width=.236\textwidth]{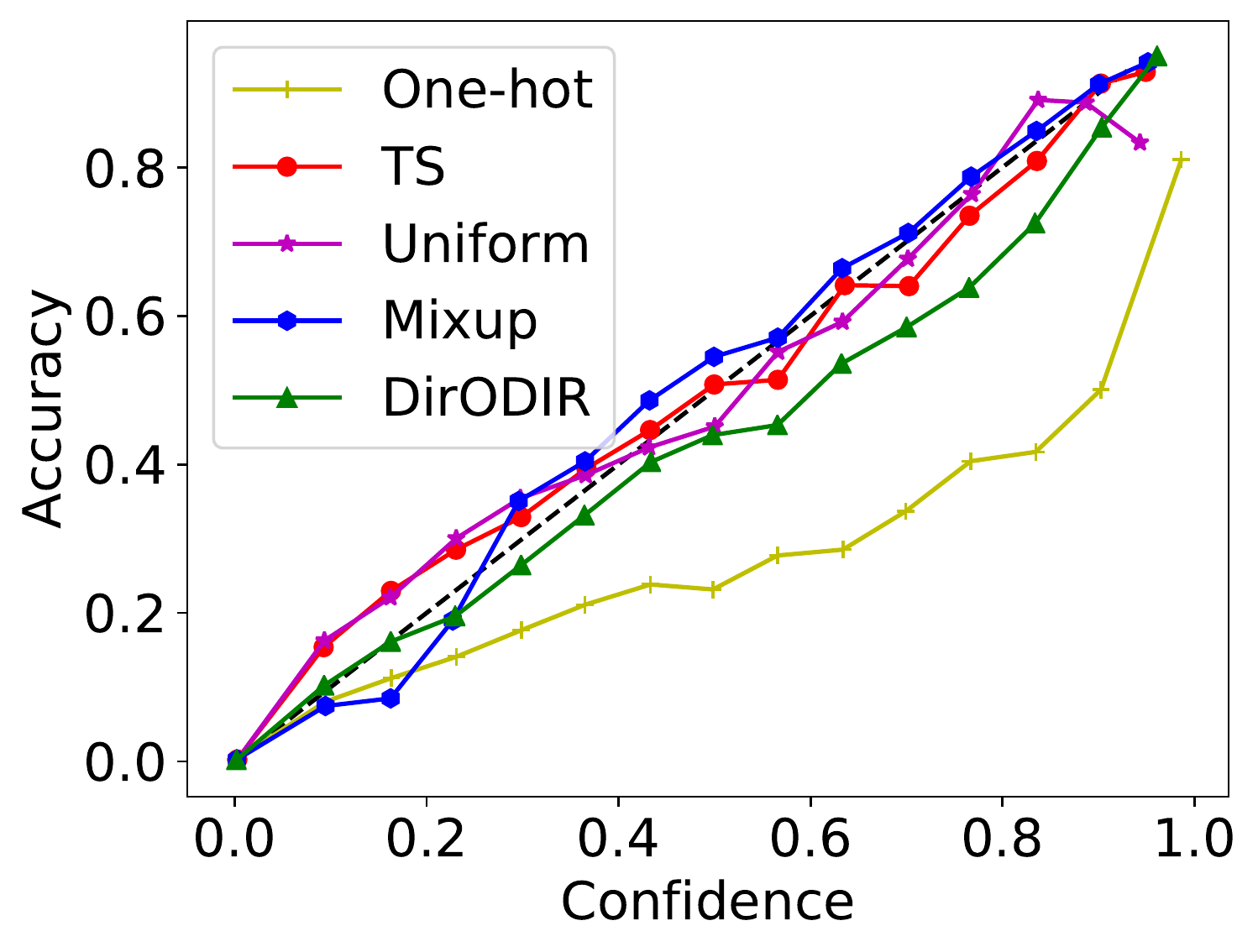}
\includegraphics[width=.24\textwidth]{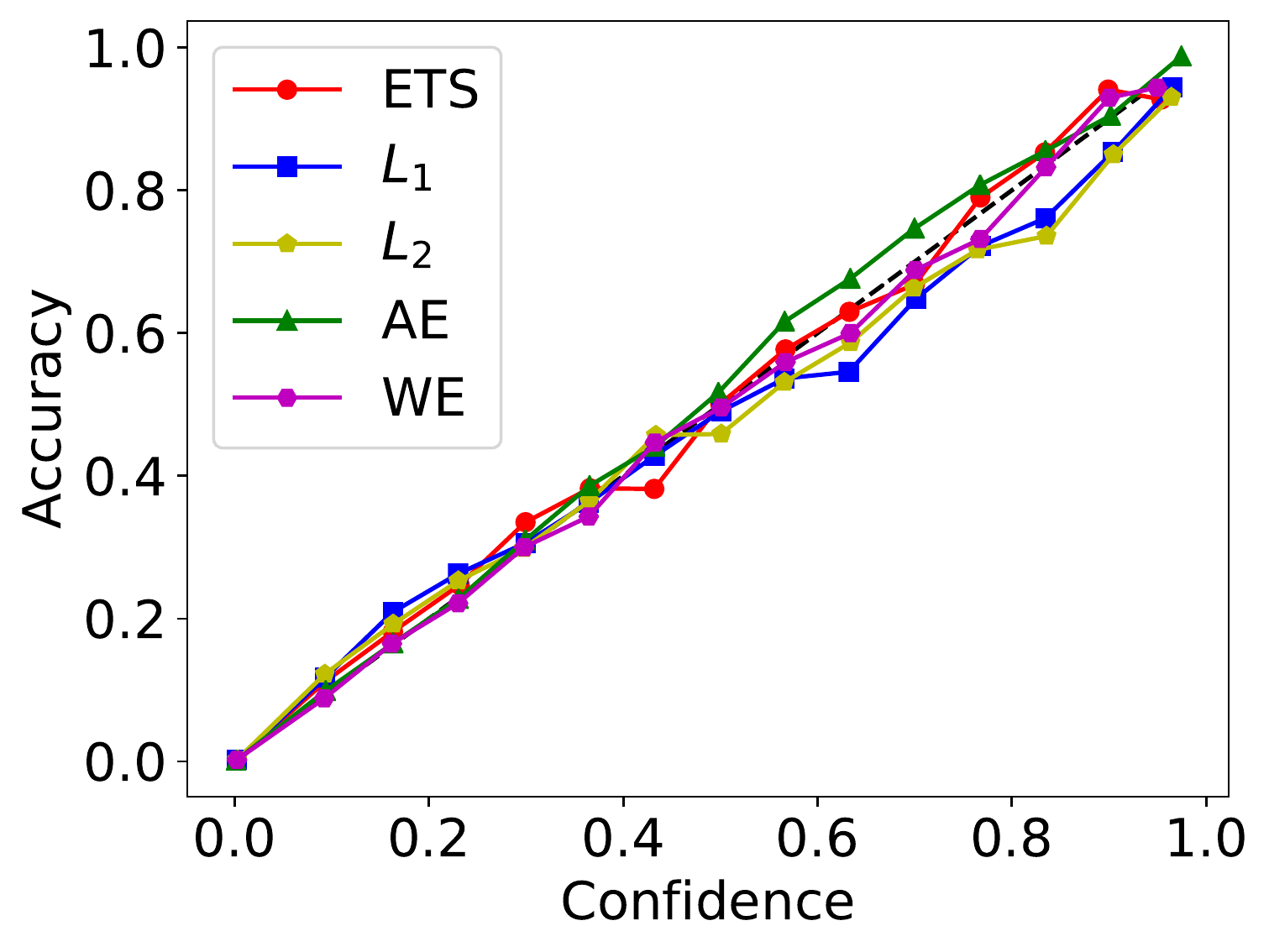}}%
\caption{Reliability diagrams on CIFAR-100 and Tiny-ImageNet dataset.\label{fig_calibration_results_diagram}}
\end{figure}

In Figure~\ref{fig_calibration_results_diagram} we show the reliability diagrams for WRN on CIFAR-100 and ResNet-34 on Tiny-ImageNet. The plots confirm our findings of the ECE results. For better readability, we divide the diagrams of the ten techniques into two figures. In all the plots, the dashed black diagonal line represents perfect calibration for which the confidence matches the accuracy. The model trained using one-hot labels is clearly overconfident since the accuracy is always below the confidence. While all other methods have a better performance, diagrams from our proposed smooth labels almost identically match the diagonal line.

\subsection{Effects of Parameters}
\label{subsection_parameters}

In this section, we perform a series of experiments on Tiny-ImageNet to explore the effects of two important hyper-parameters $\alpha$ and $\beta$.

\begin{figure}[t]
\centering
\subcaptionbox{Prediction ECE relative to $\alpha$.\label{fig_ece_alpha_pred}} {\includegraphics[width=0.29\textwidth]{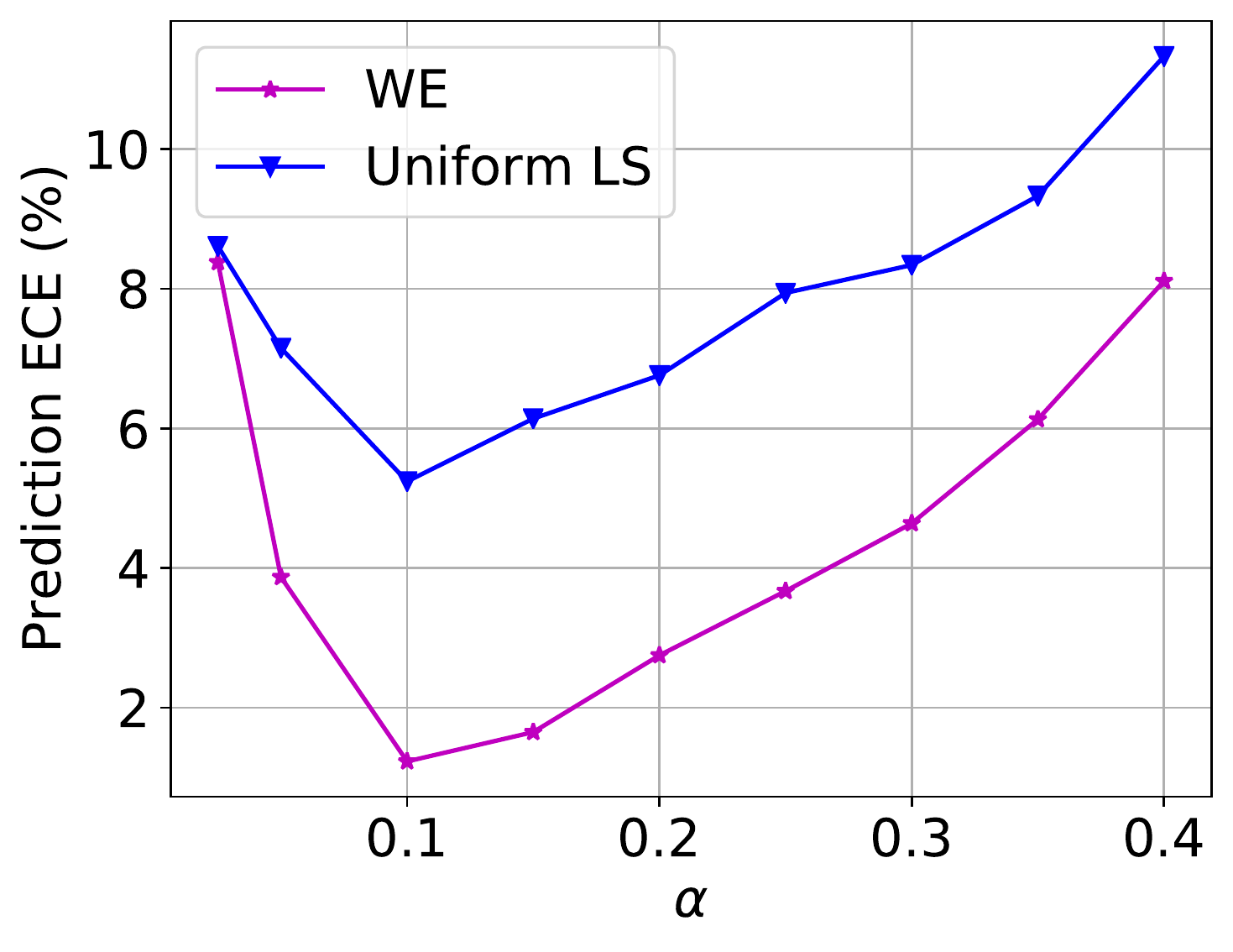}}
\hfill
\subcaptionbox{Output ECE relative to $\alpha$.\label{fig_ece_alpha_output}} {\includegraphics[width=0.29\textwidth]{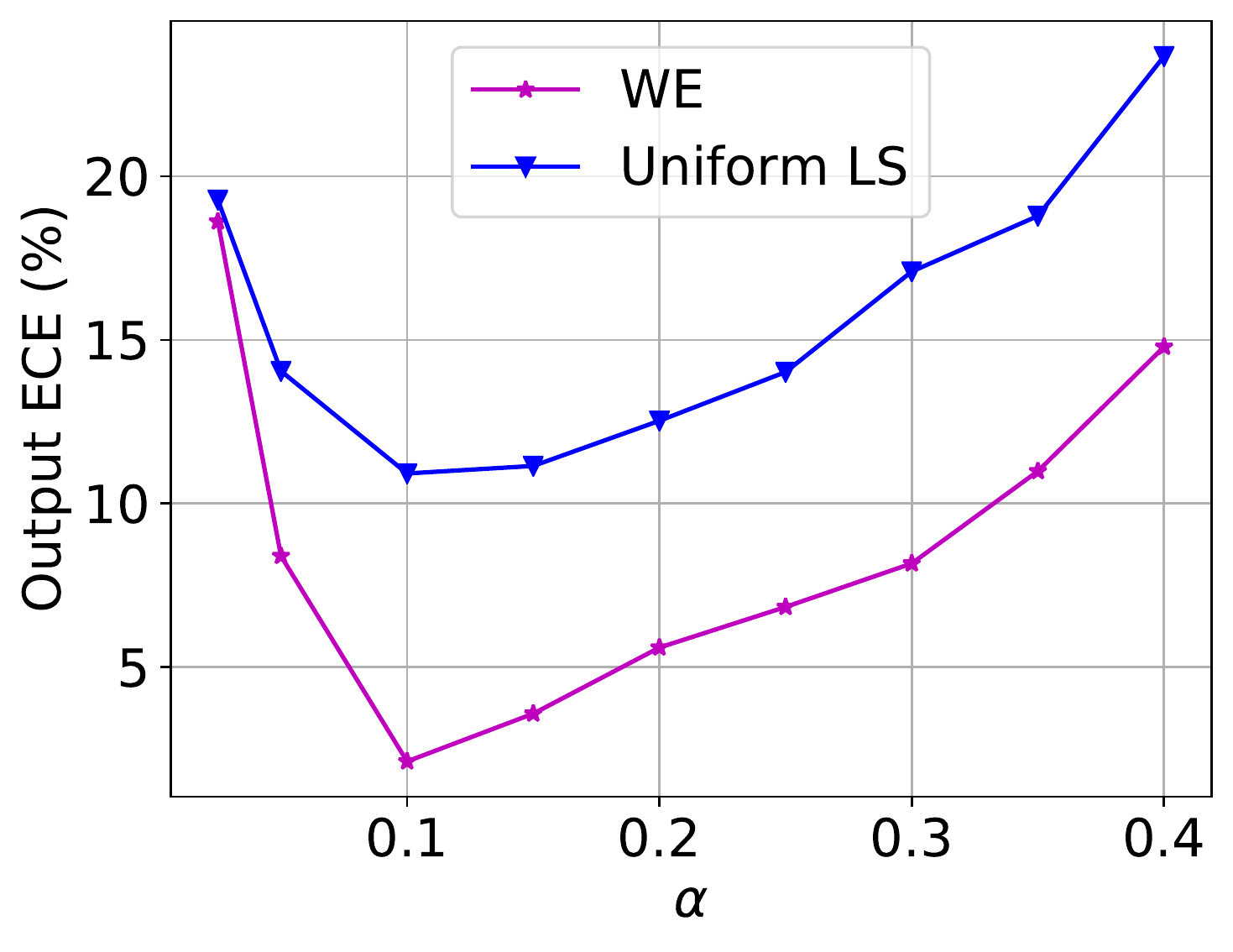}}
\hfill
\subcaptionbox{Prediction and output ECE relative to $\beta$.\label{fig_beta_plot}} {\includegraphics[width=.292\textwidth]{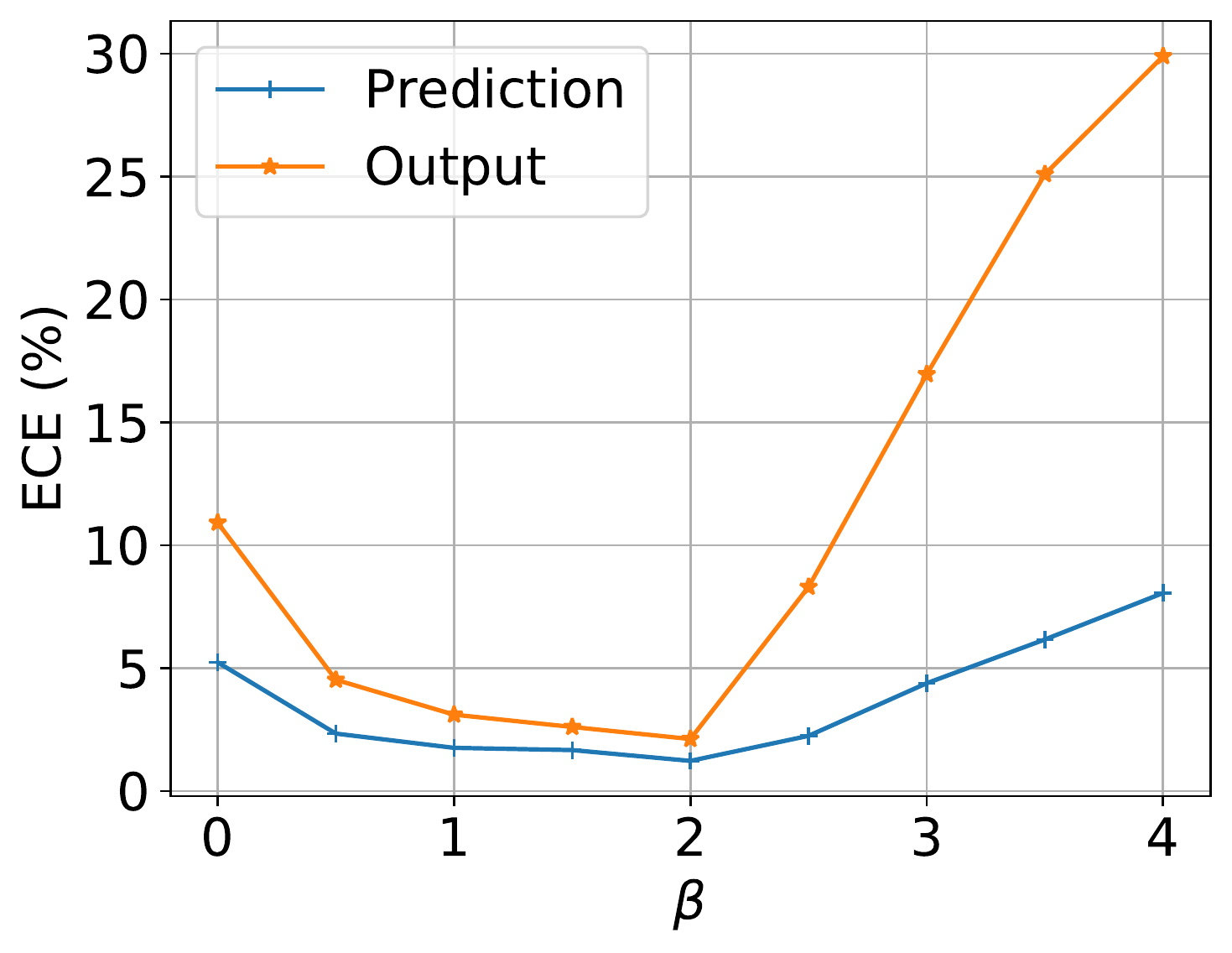}}
\caption{Effects of parameters.\label{fig_alpha_plot}}
\end{figure}

First, we test different $\alpha$ values that determine the strength of label smoothing. We compare WE to uniform LS and set $\beta=2$. The results are presented in Figure~\ref{fig_ece_alpha_pred} and~\ref{fig_ece_alpha_output}. We observe that our proposed smooth label generally outperforms uniform smooth label for all $\alpha$ values, and the best results for both methods are achieved at $\alpha=0.1$ which is the commonly value used in practice. When $\alpha$ is very small, the labels are only weakly smoothed and the model is still over-confident. When $\alpha$ becomes large, the labels are too noisy and the model is not well-calibrated because of the excessive smoothing. 

Next, we perform experiments on different $\beta$ which determines how uniform the similarity distribution is. We choose the WE model and set $\alpha=0.1$. The results are shown in Figure~\ref{fig_beta_plot}. When $\beta$ is too small, the similarity distribution is close to uniform and the class relation is not well represented, therefore the model is not well calibrated. As $\beta$ keeps increasing when it becomes too large, the softmax function will produce extreme similarity values that are concentrated in only a few classes that do not represent the optimal distribution either.

\subsection{Evaluation on Out-of-Distribution Data}
\begin{figure}[ht]
\centering
\includegraphics[width=0.3\linewidth]{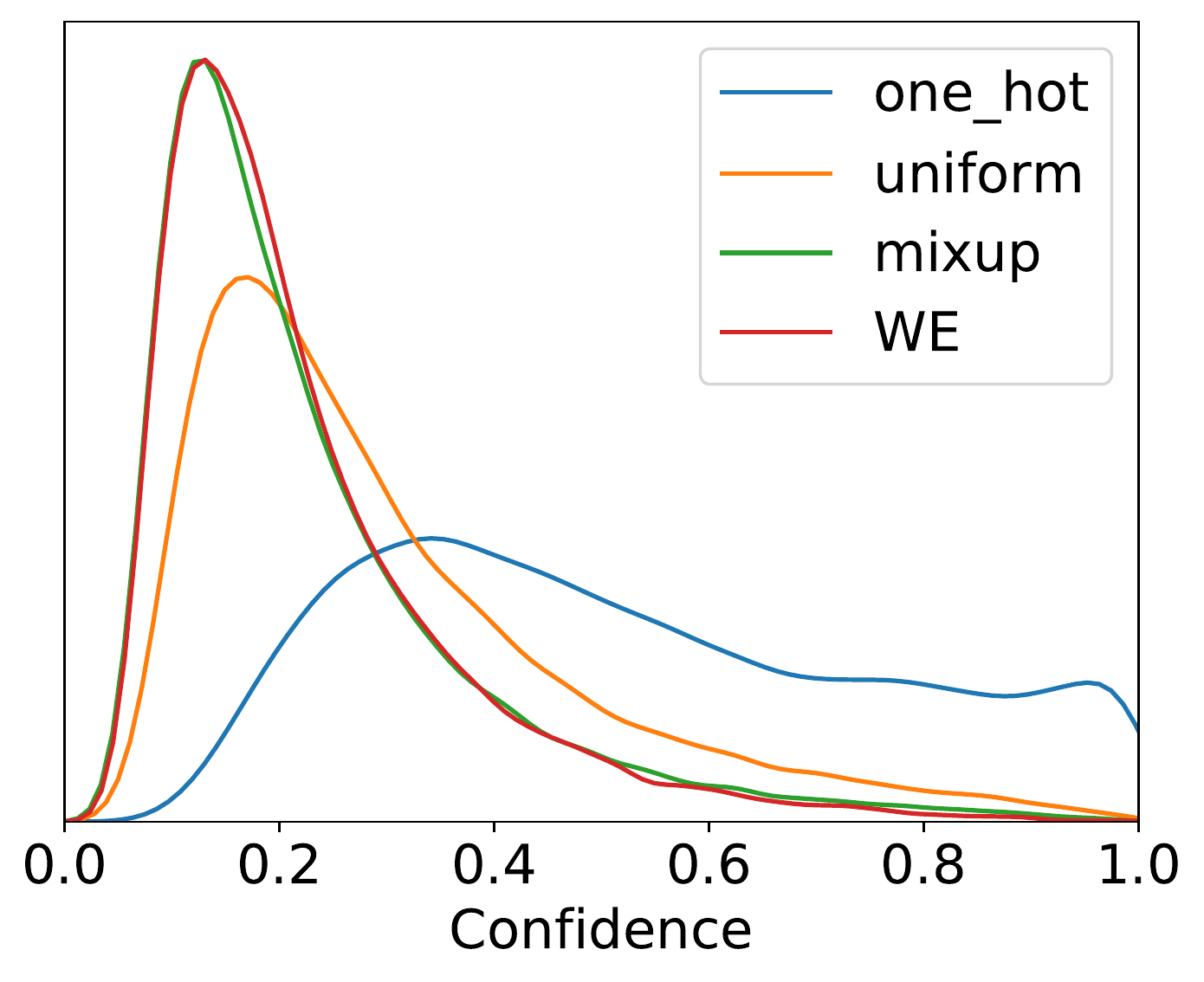}
\includegraphics[width=0.3\linewidth]{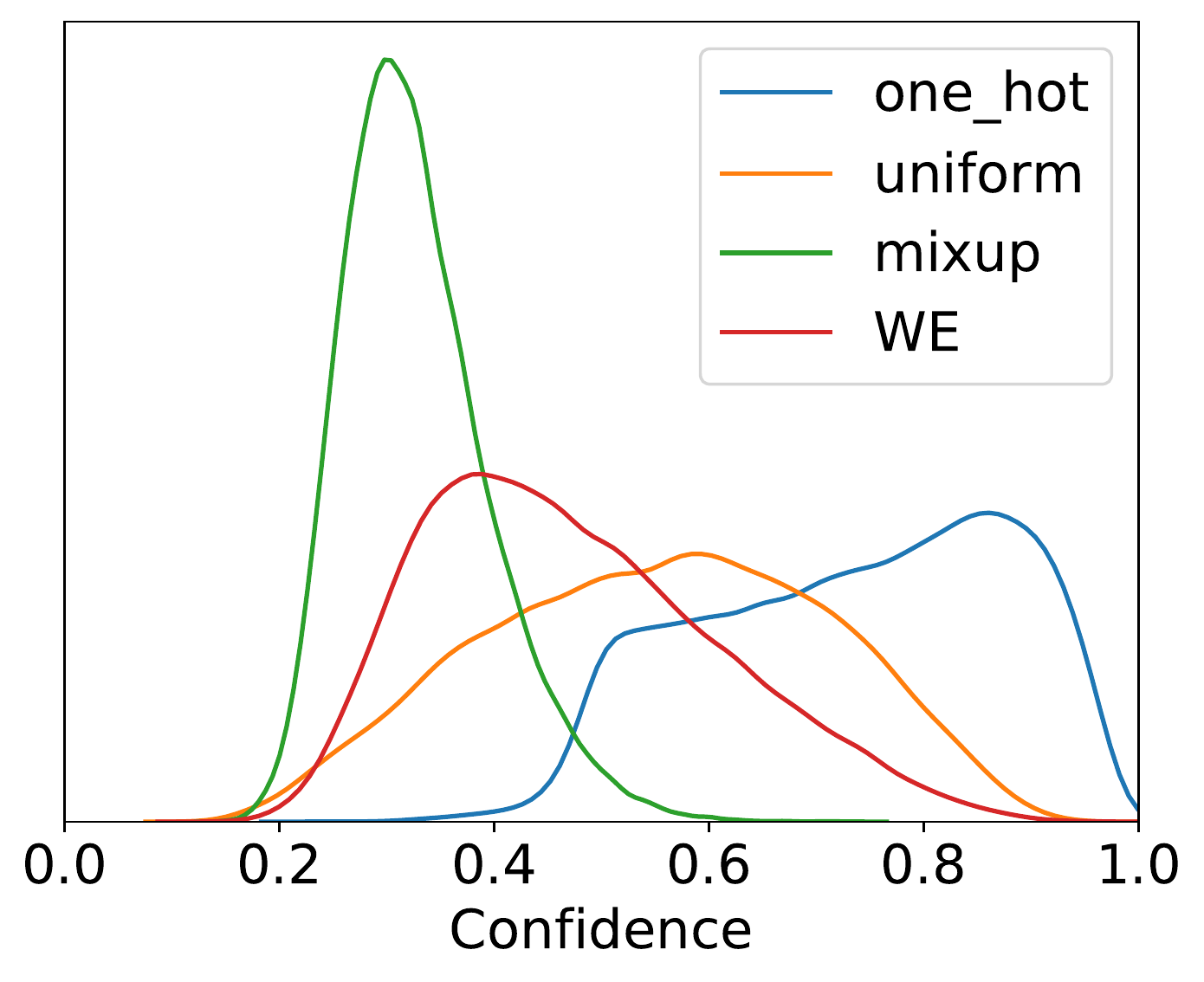}
\caption{Distribution of the prediction confidences on the validation set of Tiny-ImageNet (left) and uniform random noise (right).\label{fig_out_of_dist}}
\end{figure}

DNNs are not only overconfident on the data they are trained on but also on unseen out-of-distribution data~\cite{hendrycks2016baseline,thulasidasan2019mixup}. In this section, we evaluate the methods on two types of out-of-distribution data: another unseen dataset and random noise. We test the WRN models trained on CIFAR-100 in Section~\ref{subsection_calibration_results}. We use the validation set of Tiny-ImageNet as the unseen dataset and generate uniformly distributed random samples as the second type of out-of-distribution data. We show the distributions of the prediction confidence values in Figure~\ref{fig_out_of_dist}.

From the left plot, we can see that our method and mixup perform the best at refraining to produce high confidence on unseen out-of-distribution data. On the random noise samples, all three methods significantly outperform the one-hot training. Note that our method is based on class similarities which is not well-suited for random noise samples where the notion of similarity does not exist. Although this application is not the main focus of our method, we note that it still significantly outperforms the uniform label smoothing in both scenarios. 

\section{Conclusion}
\label{section_conclusion}
In this paper, we address the confidence calibration problem in a more holistic framework. Motivated by directly optimizing the objective of confidence calibration, we propose class-similarity based label smoothing. We adopt several similarity metrics, including those that capture feature based similarities or semantic similarity. We demonstrate through extensive experiments that our method significantly outperforms state-of-the-art techniques.

\bibliographystyle{splncs04}
\bibliography{egbib}

\end{document}